\documentclass{spieman}  

\usepackage{amsmath,amsfonts,amssymb}
\usepackage{graphicx}
\usepackage{textcomp}

\usepackage{xspace}
\newcommand{\hololens}{HoloLens\xspace}
\newcommand{\carm}{C-Arm\xspace}
\newcommand{\xray}{X-Ray\xspace}
\newcommand{\hmd}{HMD\xspace}
\newcommand{\multimarker}{multi-modality marker\xspace}
\newcommand{\artoolkit}{ARToolKit\xspace}

\newcommand{\mat}[1]{\mathbf{#1}}

\newcommand{\revision}[1]{{\color{black}#1}}
\newcommand{\revis}[1]{{\color{black}#1}}

\title{On-the-fly Augmented Reality for Orthopaedic Surgery Using a Multi-Modal 
Fiducial}

\author[a,b,*]{Sebastian Andress}
\author[c,*]{Alex Johnson, M.\,D.}
\author[a,*,$\dagger$]{Mathias Unberath}
\author[a,d,*]{Alexander Winkler}
\author[a,d]{Kevin Yu}
\author[a]{Javad Fotouhi}
\author[b]{Simon Weidert, M.\,D.}
\author[c]{Greg Osgood, M.\,D.}
\author[a,d]{Nassir Navab}
\affil[a]{Johns Hopkins University, Computer Aided Medical Procedures, 3400 N 
Charles Street, Baltimore, MD, USA}
\affil[b]{Ludwig-Maximilians-Universit\"at M\"unchen, Klinik f\"ur Allgemeine, 
Unfall- und Wiederherstellungschirurgie, Nu{\ss}baumstra{\ss}e 20, Munich, 
Germany}
\affil[c]{Johns Hopkins Hospital, Department of Orthopaedic Surgery, 1800 
Orleans Street, Baltimore, MD, USA}
\affil[d]{Technische Universit\"{a}t M\"{u}nchen, Computer Aided Medical 
Procedures, Boltzmannstra{\ss}e 3, Munich, Germany}

\begin{document} 
	\maketitle
	
	\begin{abstract}
		Fluoroscopic \xray guidance is a cornerstone for percutaneous 
		orthopaedic surgical procedures. However, two-dimensional observations 
		of the three-dimensional anatomy suffer from the effects of projective 
		simplification. Consequently, many \xray images from various 
		orientations need to be acquired for the surgeon to accurately assess 
		the spatial relations between the patient's anatomy and the surgical 
		tools.\\
		In this paper, we present an on-the-fly surgical support system that 
		provides guidance using augmented reality and can be used in 
		quasi-unprepared operating rooms. The proposed system builds upon a 
		\multimarker and simultaneous localization and mapping technique to 
		co-calibrate an optical see-through head mounted display to a \carm 
		fluoroscopy system. 
		Then, annotations on the 2D \xray images can be rendered as virtual 
		objects in 3D providing surgical guidance.\\
		We quantitatively evaluate the components of the proposed system, and 
		finally, design a feasibility study on a semi-anthropomorphic phantom. 
		The accuracy of our system was comparable to the traditional 
		image-guided technique while substantially reducing the number of 
		acquired \xray images as well as procedure time.\\
		Our promising results encourage further research on the interaction 
		between virtual and real objects, that we believe will directly benefit 
		the proposed method. Further, we would like to explore the capabilities 
		of our on-the-fly augmented reality support system in a larger study 
		directed towards common orthopaedic interventions.
	\end{abstract}
	
	
	{\noindent \footnotesize\textbf{$\dagger$}  Please send correspondence to 
	\linkable{unberath@jhu.edu}.}
	
	{\noindent \footnotesize\textbf{*} These authors have contributed equally 
	and are listed in alphabetical order.}
	
	\begin{spacing}{1.3}   
		
		\section{Introduction}
		\label{sec:intro} 
		
		\revision{
			Minimally-invasive and percutaneous procedures with small incisions 
			are an ongoing trend in orthopedic 
			surgery~\cite{gay1992percutaneous}. As anatomic structures and 
			location of tools and implants are not directly visible to the 
			human eye, intra-operative imaging is needed for safe and effective 
			procedures. The standard imaging modality still is fluoroscopic 
			imaging by a \carm device, displaying 2-D X-ray projection images 
			to the surgeon on a separate monitor.
			The most demanding tasks using fluoroscopy are those involving the 
			precise placement of tools or implants. Due to the 2-D projection, 
			the position of an instrument’s tip for instance on the axis 
			perpendicular to the image plane cannot be safely determined. To 
			compensate for this, additional images have to be acquired from 
			another angle, ideally 90$^\circ$ to create two-planar images that 
			allow the surgeon to estimate the true 3-D position of any 
			radiopaque structure within the field of 
			view~\cite{boszczyk2006fluoroscopic}. 
			The mental projection of 2-D images onto a 3-D world often is 
			counterintuitive and error-prone due to projective simplification 
			and high mental work-load. The result often is repetitive X-ray 
			imaging to control the procedure, increasing the radiation dose for 
			the patient but especially for the surgeon who is often standing 
			very close to the patient with his hands close to the beam 
			\cite{Kirousis2009,Sugarman1988}. Another problem is that 
			fluoroscopic imaging is non-continuous but rather provides single 
			snapshots, leaving the surgeon ''blind'' in-between image 
			acquisitions. Both, projection errors and non-continuous imaging 
			can lead to misplacement of tools or implants, potentially leading 
			to tissue damage with consequences ranging from hematoma and 
			bleeding to nerve damage or joint 
			destruction~\cite{robb2003guidewire,zionts1991transient}. The 
			potential results of this are poor outcomes and revision rates.
			
			Computer aided surgery (CAS), also known as computer navigation has 
			formerly been introduced to offer the surgeon a more powerful 
			visualization of his actions within the 3D space, continuously 
			displaying tracked instruments or implants within the patient’s 
			anatomy mostly using pre-existing CT or MRI datasets. Those systems 
			are used in a wide range of procedures ranging from neurosurgery to 
			orthopedic surgery. All of those systems require an initial 
			registration process and calibration of specialized tools. In most 
			cases, the patient as well as the instruments are tracked by 
			optical markers attached to them visible to infrared cameras. An 
			abundance of literature exists on the benefits of these systems on 
			accuracy and safety~\cite{gras2012screw}. But other tracking 
			solutions such as electromagnetic tracking have also shown the 
			ability to significantly reduce radiation without substantially 
			increasing surgical times~\cite{maqungo2014distal}, while 
			introducing further requirements as an electromagnetically shielded 
			operating field.
			Despite the positive effects described, CAS still is far from 
			dominating the operating room. The reasons for this being not only 
			the high cost of investment but also the added surgery time caused 
			by technical setup and additional workflow steps such as repeated 
			registration. Studies saw an increase in total procedure time up to 
			65 minutes without clearly improving clinical 
			outcomes~\cite{Hawi2014,Wilharm2013,Hamming_Automatic_2009,Doke2015,kraus2015computer}.
			 Furthermore, those systems are often associated with a long 
			learning curve until an efficient workflow is established and the 
			team learned how to prevent or deal with technical difficulties 
			such as registration errors and line-of-sight problems while 
			tracking~\cite{Victor2004}. 
			Another reason clearly is the distraction caused by displaying 
			visualization on a separate monitor far from the real action and in 
			a totally different coordinate system as the real world, limiting 
			ergonomics and usability especially for the new 
			user~\cite{qian2017technical}. Therefore, there is much potential 
			for improvement and new technologies that offer a more intuitive 
			approach and facilitate percutaneous surgery without introducing a 
			complicated setup or requiring major changes to the operative 
			workflow. 
			
			Augmented Reality (AR) is a technology that promises to integrate 
			computer guidance into the surgical workflow in a very intuitive 
			way by providing visual guidance directly related to the anatomical 
			target area inside the patient in front of the surgeon’s eyes. 
			Current CAS systems usually display information on 2-D monitors 
			mounted on carts or the ceiling of the 
			theater~\cite{navab2010camera,fotouhi2017plan}. As the relation 
			between the monitor and the patient or the surgeon is unknown to 
			the system, it displays a view onto the anatomy as well as tool 
			trajectories that are unrelated to the real perspective of the 
			surgeon. 
			AR systems to support orthopedic surgery have been demonstrated and 
			evaluated already. One of the most studied designs consists of a 
			\carm with a calibrated video camera attached to it, thus being 
			able to augment the live video image with fluoroscopic images in 
			precise overlay~\cite{navab2010camera}. Its most recent variation 
			relies on an RGBD camera rigidly mounted on the \carm detector and 
			calibrated to an intra-operative cone-beam CT (CBCT) scan. It has 
			the ability to simultaneously render multiple digitally 
			reconstructed radiographs at different viewing angles overlaid with 
			a 3-D cloud of points from the RGBD camera showing the surgeon's 
			hands and tools \cite{lee2016calibration,fischer2016preclinical}. 
			Despite promising performance that includes remarkable reductions 
			in both surgery time and dose without substantial changes to the 
			traditional workflow, the system requires an intra-operative CBCT 
			scan which is only available on high-end \carm systems and adds 
			operation time and dose. Replacing the CBCT with 2-D/3-D 
			registration of pre-operative CT to interventional fluoroscopy 
			mitigates aforementioned challenges and makes the system more 
			usable for many orthopeadic surgical 
			applications~\cite{fotouhi2017pose, tucker2018towards}. The use of 
			\hmd for AR in orthopaedic surgery is suggested in the literature 
			for guiding the placement of percutaneous sacroiliac screws in 
			pelvic fractures~\cite{wang2016precision}. This approach relies on 
			external navigation systems to track the drill, pelvis, and \hmd.
			
			Within this manuscript, we propose an easy-to-use guidance system 
			with the specific aim of eliminating potential roadblocks to its 
			use regarding system setup, change in workflow, or cost. The system 
			is applicable to most fluoroscopy-guided orthopeadic surgeries 
			without the need for 3D pre- or intra-operative imaging, and 
			provides support for surgeon's actions through an AR environment 
			based on optical see-through \hmd that is calibrated to the \carm 
			system. The proposed solution eliminates the need for external 
			navigation hardware, as well as the pre-operative calibration of 
			the sensors.
			It allows visualizing the path to anatomical landmarks annotated in 
			\xray images in 3-D directly on the patient. Calibration of 
			intra-operative fluoroscopy imaging to the AR environment is 
			achieved on-the-fly using a mixed-modality fiducial that is imaged 
			simultaneously by the \hmd and the \carm system. Therefore, the 
			proposed system effectively avoids the use of dedicated but 
			impractical optical or electromagnetic tracking solutions with 
			2-D/3-D registration, complicated setup or use of which is 
			associated with the most substantial disruptions to the surgical 
			workflow.
			The aim of this study is to describe the calibration of the system 
			and to determine its performance in a set of experiments carried 
			out by expert surgeons. The first experiment will determine 
			accuracy and precision when using a technical phantom consisting of 
			spherical radiopaque target points that are surrounded by 
			soft-tissue. The second experiment mimics a typical step in many 
			surgical procedures, percutaneously placing a K-wire onto a 
			specific location of the patient’s anatomy. In our case, the task 
			will be finding the entry-point for an intra-medullary nail at the 
			tip of the greater trochanter. This task is typical for most 
			percutaneous orthopedic procedures such as pedicle screw, lag screw 
			placement, or interlocking of nails.
			
			The manuscript is outlined as follows.
			In sec.~\ref{sec:intro} we discussed problems of the classic 
			fluoroscopic guidance technique during minimal invasive surgeries. 
			A concrete example describes the need for another guidance 
			technique. Current solutions in different research fields are 
			addressed, leading to the gap of a simple on-the-fly guidance 
			technique for surgeries. 
			Sec.~\ref{sec:method} provides an overview of the presented 
			solution. In the following, the used devices and their spatial 
			transformations are explained.
			A \multimarker is introduced for the registration of the system 
			using visual marker tracking. We then discuss the \carm and its 
			integration with the system. Next, we describe how \hmd devices are 
			used together with the \carm scanner to enable real-time AR. 
			Four experiments are then presented to evaluate the accuracy and 
			precision of each individual component, followed by a summary of 
			the results in Sec.~\ref{sec:results}. Lastly, we discuss the 
			experimental results and provide a summary of the proposed AR 
			solutions in Sec.~\ref{sec:discussion} and~\ref{sec:conclusion}, 
			respectively.
		}

		\section{Method}
		\label{sec:method}

		\subsection{System Overview}
		\label{subsec:overview}
\begin{figure}
			\includegraphics[width=\linewidth]{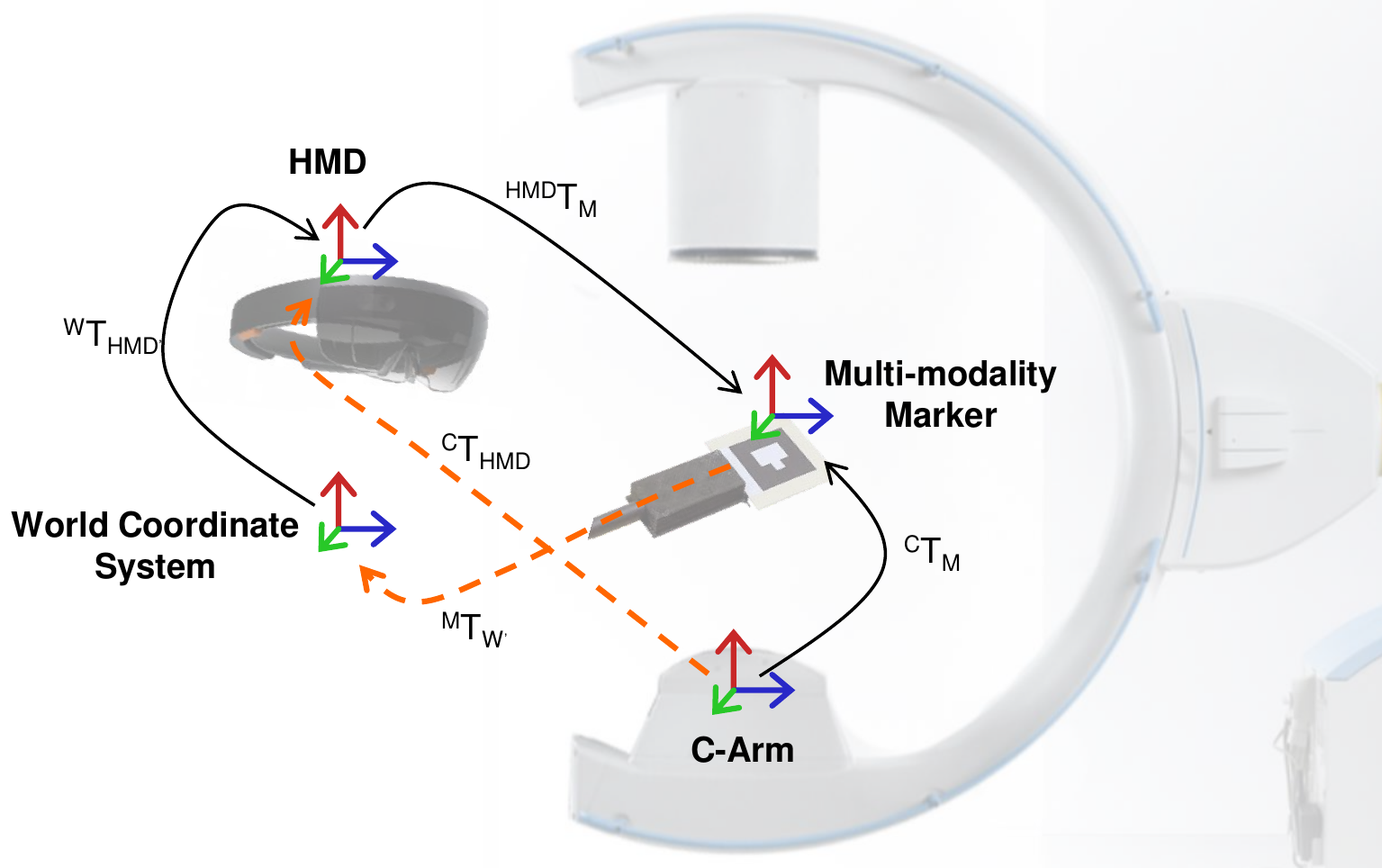}
			\centering
			\caption{Spatial transformations for the on-the-fly AR solution.}
			\label{fig:transformations}
		\end{figure}

		The proposed system comprises three components that must exhibit 
		certain characteristics to enable on-the-fly AR guidance: a 
		mixed-modality fiducial, a \carm \xray imaging system, and an optical 
		see-through \hmd. \revision{We will use the following notation of 
		expressing transformations: The transformation 
		\(^\text{A}\mat{T}_\text{B}\) is defined as the transformation from 
		coordinate system A to coordinate system B. This notation enables to 
		concatenate transformations easily as in \(^\text{A}\mat{T}_\text{C} = 
		^\text{B}\mat{T}_\text{C} ~ ^\text{A}\mat{T}_\text{B}\).} Based on 
		these components, the spatial relations that need to be estimated in 
		order to enable real-time AR guidance are shown in 
		Fig.~\ref{fig:transformations}. Put concisely, we are interested in 
		recovering the transformation $^\text{C}\mat{T}_\text{HMD}(t)$ that 
		propagates information from the C-arm to \hmd coordinate system while 
		the surgeon moves over time $t$. To this end, we need to estimate the 
		following transformations:
		\revision{
			\begin{itemize}
				\item $^\text{C}\mat{T}_\text{M}$: Extrinsic calibration of the 
				\carm to the \multimarker domain, obtained from the \xray image 
				using \artoolkit (see Section~\ref{subsec:marker}).
				\item $^\text{HMD}\mat{T}_\text{M}$: Transformation describing 
				the relation between the \hmd and the \multimarker coordinate 
				system, estimated from the RGB image acquired by the HMD using 
				\artoolkit (Section~\ref{subsec:marker}).
				\item $^\text{W}\mat{T}_\text{HMD}$: \revis{The \hmd is capable 
				of establishing a map of its surroundings with arbitrary origin 
				while localizing itself therein. $^\text{W}\mat{T}_\text{HMD}$ 
				then describes the pose of the \hmd within this so-called world 
				coordinate system. In practice, it is computed using 
				vision-based tracking algorithms such as simultaneous 
				localization and mapping~\cite{durrant2006slam} 
				(Section~\ref{subsec:hmd}).} 
				\item $^\text{M}\mat{T}_\text{W}$: Describes the mapping from 
				the \multimarker to the world coordinate system. It is 
				estimated using $^\text{W}\mat{T}_\text{HMD}$ and 
				$^\text{HMD}\mat{T}_\text{M}$ in the calibration phase (see 
				Section~\ref{subsec:hmd}).
			\end{itemize}
		}
		Once these relations are known, annotations in an intra-operatively 
		acquired \xray image can be propagated to and visualized by the \hmd 
		which provides support for placement of wires and screws in orthopaedic 
		interventions. The transformation needed is given by:
		\begin{equation}
		\label{eq:spatialRelation}
		^\text{C}\mat{T}_\text{HMD}(t) = ^\text{W}\mat{T}_\text{HMD}(t) ~ 
		\underbrace{\bigg(\ ^\text{W}\mat{T}_\text{HMD}^{-1}(t_0) ~ 
		^\text{HMD}\mat{T}_\text{M}^{-1}(t_0)  \bigg) ~ 
		^\text{C}\mat{T}_\text{M}(t_0)}_{^\text{C}\mat{T}_\text{W}}\,,
		\end{equation}
		where $t_0$ denotes the time of calibration, e.\,g. directly after 
		repositioning of the \carm, suggesting that $^\text{C}\mat{T}_\text{W}$ 
		is constant as long as the \carm remains in place. For brevity of 
		notation, we will omit the time dependence of the transformations 
		whenever they are clear or unimportant.\\
		We provide detailed information on the system components and how they 
		are used to estimate aforementioned transformations in the following 
		sections.
		
		\subsection{Multi-modality Marker}
		\label{subsec:marker} 
\begin{figure}
			\centering
			\includegraphics[width=\linewidth]{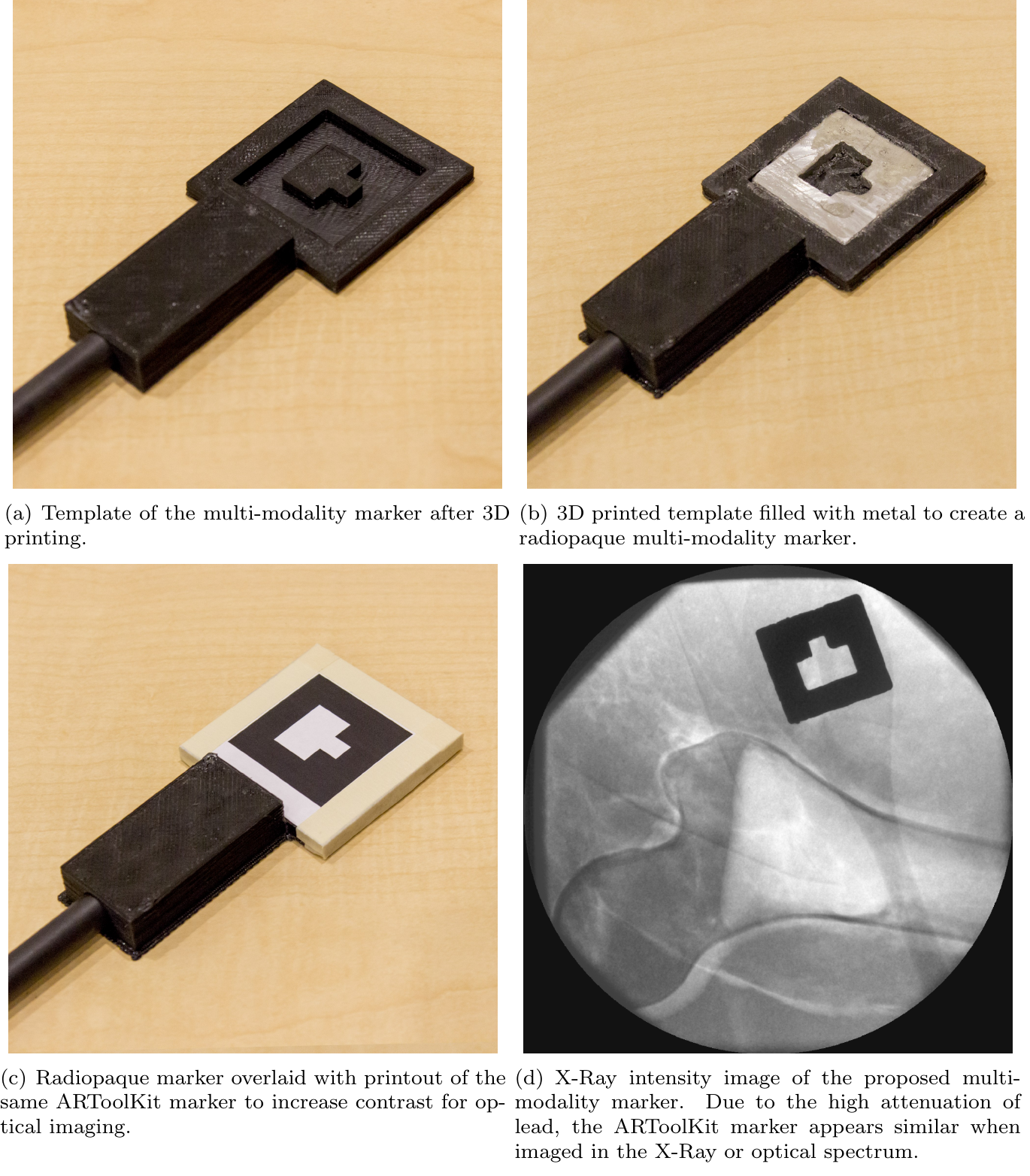}
			\caption{Steps in the creation of the \multimarker. The 3-D printed 
			template serves as a housing for the marker and is rigidly attached 
			to a carbon fiber rod such that the marker can be safely introduced 
			into the \xray field of view.}
			\label{fig:marker}
		\end{figure}

		The key component of the proposed system is a \multimarker that can be 
		detected using \carm as well as the \hmd using \xray and RGB imaging 
		devices, respectively. As the shape and size of the \multimarker is 
		precisely known in 3-D, estimation of both transforms 
		$^\text{C}\mat{T}_\text{M}$ and $^\text{HMD}\mat{T}_\text{M}$ is 
		possible in a straightforward manner if the marker can be detected in 
		the 2-D images. To this end, we rely on the well-known \artoolkit for 
		marker detection and calibration~\cite{kato1999marker}, and design our 
		\multimarker accordingly.\\
		The marker needs to be well discernible when imaged using the optical 
		and \xray spectrum. To this end, we 3-D print the template of a 
		conventional \artoolkit marker as shown in Fig.~\ref{fig:marker}(a) 
		that serves as the housing for the \multimarker. Then, we machined a 
		metal inlay (solder wire 60\textbackslash40 Sn\textbackslash Pb) that 
		strongly attenuates \xray radiation, see Fig.~\ref{fig:marker}(b). 
		After covering the metal with a paper printout of the same \artoolkit 
		marker as shown in Fig.~\ref{fig:marker}(c), the marker is equally well 
		visible in the \xray spectrum as well as RGB images due to the high 
		attenuation of lead as can be seen in Fig.~\ref{fig:marker}(d). This is 
		very convenient, as the same detection and calibration pipeline readily 
		provided by \artoolkit can be used for both images.\\
		It is worth mentioning that the underlying vision-based tracking method 
		in \artoolkit is designed for reflection and not for transmission 
		imaging which can be problematic in two ways. First, \artoolkit assumes 
		2-D markers suggesting that the metal inlay must be sufficiently thin 
		in order not to violate this assumption.  Second, a printed marker 
		imaged with an RGB camera perfectly occludes the scene behind it and 
		is, thus, very well visible. For transmission imaging, however, this is 
		not necessarily the case as all structures along a given ray contribute 
		to the intensity at the corresponding detector pixel. If other strong 
		edges are present close to this hybrid marker, detection and hence 
		calibration may fail. To address both problems simultaneously we use 
		digital subtraction, a concept that is well known from 
		angiography~\cite{chilcote1981digital,unberath2016virtual}. We acquire 
		two \xray images using the same acquisition parameters and \carm pose 
		both with and without the \multimarker introduced into the 
		\xray beam. Logarithmic subtraction then yields an image that, ideally, 
		only shows the \multimarker and lends itself well to marker detection 
		and calibration using the \artoolkit pipeline. Moreover, subtraction 
		imaging allows for the use of very thin metal inlays as subtraction 
		artificially increases the contrast achieved by attenuation only. While 
		the subtraction image is used for processing, the surgeon is shown the 
		fluoroscopy image without any \multimarker obstructing the scene.

		\subsection{\carm Fluoroscopy System}
		\label{subsec:carm}
		The proposed system has the substantial advantage that, in contrast to 
		many previous 
		systems~\cite{fischer2016preclinical,fotouhi2016interventional}, it 
		does not require any modifications to commercially available \carm 
		fluoroscopy systems. The only requirement is that images acquired 
		during the intervention can be accessed directly such that geometric 
		calibration is possible. Within this work, we use a Siemens ARCADIS 
		Orbic 3D (Siemens Healthcare GmbH, Forchheim, Germany) to acquire 
		fluoroscopy images and a frame grabber (Epiphan Systems Inc, Palo Alto, 
		CA) paired with a streaming server~\cite{qian2017technical} to send 
		them via a wireless local network to the \hmd.\\
		While extrinsic calibration of the \carm system is possible using the 
		\multimarker as detailed in Sec.~\ref{subsec:marker}, the intrinsic 
		parameters of the \carm, potentially at multiple poses, are estimated 
		in a one-time offline calibration, e.\,g. as described by Fotouhi et 
		al.~\cite{fotouhi2017can} using a radiopaque checkerboard.\\
		Once the extrinsic and intrinsic parameters are determined, the 3-D 
		source and detector pixel positions can be computed in the coordinate 
		system of the \multimarker. This is beneficial, as simple point 
		annotations on the fluoroscopy image now map to lines in 3-D space that 
		represent the \xray beam emerging from the source to the respective 
		detector pixel. These objects, however, cannot yet be visualized at a 
		meaningful position as the spatial relation of the \carm to the \hmd is 
		unknown. The \multimarker enabling calibration must be imaged 
		simultaneously by the \carm system and the RGB camera on the \hmd to 
		enable meaningful visualization in an AR environment. This process will 
		be discussed in greater detail below.
		
		\begin{figure}
			\centering
			\includegraphics[width=\linewidth]{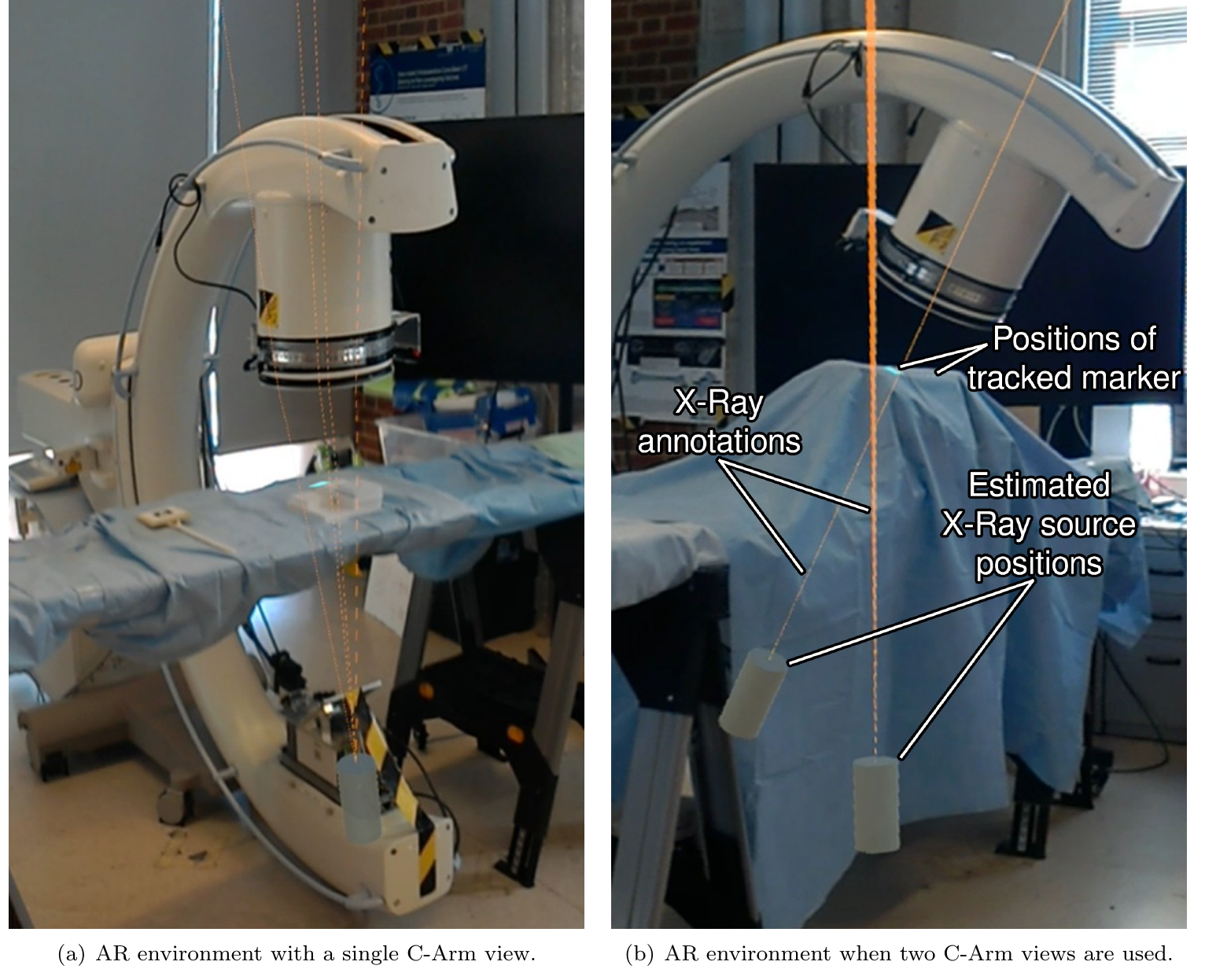}
			%
			%
			\caption{Source position of the \carm shown as a cylinder and 
			virtual lines that arise from annotations in the fluoroscopy image.}
			\label{fig:xraySource}
		\end{figure}

		\subsection{Optical See-through \hmd and the World Coordinate System}
		\label{subsec:hmd}
		The optical see-through \hmd is an essential component of the proposed 
		system as it needs to recover its pose with respect to the world 
		coordinate system at all times, acquire and process optical images of 
		the \multimarker, allow for interaction of the surgeon with the 
		supplied \xray image, combine and process the information provided by 
		the surgeon and the \carm, and provide real-time AR visualization for 
		guidance. Within this work, we rely on the Microsoft \hololens 
		(Microsoft Corporation, Redmond, WA) as the optical see-through \hmd as 
		its performance compared favorably to other commercially available 
		devices~\cite{qian2017comparison}. 
		
		\paragraph{Pose Estimation}
		Similar to the pose estimation for the \carm, we first seek to estimate 
		the pose of the \hmd with respect to the \multimarker 
		$^\text{HMD}\mat{T}_\text{M}$. In order to allow for calibration of the 
		\carm to the \hmd, the images of the marker used to retrieve 
		$^\text{C}\mat{T}_\text{M}$ and $^\text{HMD}\mat{T}_\text{M}$ for the 
		\carm and the \hmd, respectively, must be acquired with the marker at 
		the same position. If the \multimarker is hand-held, the images should 
		ideally be acquired at the same time $t_0$. The \hololens is equipped 
		with an RGB camera that we use to acquire an optical image of the 
		\multimarker and estimate $^\text{HMD}\mat{T}_\text{M}$ using 
		\artoolkit as described in Sec.~\ref{subsec:marker}.\\
		In principle, these two transformations are sufficient for AR 
		visualization but the system would not be appropriate: if the surgeon 
		wearing the \hmd moves, the spatial relation 
		$^\text{HMD}\mat{T}_\text{M}$ changes. As limiting the surgeons 
		movements is not feasible, updating $^\text{HMD}\mat{T}_\text{M}(t)$ 
		over time may seem like an alternative but is impracticable as it would 
		require the \multimarker to remain at the same position, potentially 
		close to the operating field. While updating 
		$^\text{HMD}\mat{T}_\text{M}(t)$ over time seems complicated, 
		recovering $^\text{HMD}\mat{T}_\text{W}(t)$, the pose of the \hmd with 
		respect to the world coordinate system, is readily available from the 
		\hololens \revis{\hmd and is estimated using a proprietary algorithm 
		based on concepts similar to simultaneous localization and mapping 
		(SLAM)~\cite{durrant2006slam,Microsoft2017,kress2017hololens}.} 
		Consequently, rather than directly calibrating the \carm to the \hmd, 
		we calibrate the \carm to the world coordinate system (in the \hololens 
		community sometimes referred to as world anchor or spatial map) to 
		retrieve $^\text{C}\mat{T}_\text{W}$ that is constant if the \carm is 
		not repositioned.
		
		\paragraph{User Interface and AR Visualization}
		In order to use the system for guidance, key points must be identified 
		in the \xray images. 
		Intra-operative fluoroscopy images are streamed from the \carm to the 
		\hmd and visualized using a \emph{virtual monitor} as described in 
		greater detail in Qian et al.~\cite{qian2017technical}. The surgeon can 
		annotate anatomical landmarks in the \xray image by hovering the 
		\hololens cursor over the structure and performing the \emph{air tap} 
		gesture. In 3-D space, these points must lie on the line connecting the 
		\carm source position and the detector point that can be visualized to 
		guide the surgeon using the spatial relation in 
		Eq.~\ref{eq:spatialRelation}. An exemplary scene of the proposed AR 
		environment is provided in Fig.~\ref{fig:xraySource}. Guidance rays are 
		visualized as semi-transparent lines with a thickness of \(1\)\,mm 
		while the \carm source position is displayed as a cylinder. The 
		association from annotated landmarks in the \xray image to 3-D virtual 
		lines is achieved via color coding.\\
		It is worth mentioning that the proposed system allows for the use of 
		two or more \carm poses simultaneously. When two views are used, the 
		same anatomical landmark can be annotated in both fluoroscopy images 
		allowing for stereo reconstruction of the landmark's 3-D 
		position~\cite{hartley2003multiple}. In this case, a virtual sphere is 
		shown in the AR environment at the position of the triangulated 3-D 
		point, shown in Fig.~\ref{fig:xraySource}(b). Furthermore, the 
		interaction allows for the selection of two points in the same \xray 
		image that define a line. This line is then visualized as a plane in 
		the AR environment. An additional line in a second \xray image can be 
		annotated resulting in a second plane. The intersection of these two 
		planes in the AR space can be visualized by the surgeon and followed as 
		a trajectory.

		\subsection{Integration with the Surgical Workflow}
		\label{subsec:integration}
		
		As motivated in section \ref{sec:intro}, one of the main goals of this 
		study was to create an easy on-the-fly guidance system. The simple 
		setup proposed here is enabled by the \multimarker and, more 
		substantially, by the capabilities of the \hololens.\\
		Configuring the system in a new operating room requires access to the 
		\carm fluoroscopy images and setup of a local wireless data transfer 
		network. Once the \hmd is connected to the \carm, only very few steps 
		for obtaining AR guidance are needed. For each \carm pose, the surgeon 
		has to:
		\begin{enumerate}
			\item Position the \carm using the integrated laser cross-hair such 
			that the target anatomy will be visible in fluoroscopy.
			\item Introduce the \multimarker in the \carm field of view and 
			also visible in the RGB camera of the \hmd. If the fiducial is 
			recognized by the \hmd, an overlay will be shown. Turning the head 
			such that the marker is visible to the eye in straight gaze is 
			usually sufficient to achieve marker detection.
			\item Calibrate the system by use of a voice command (''Lock'') and 
			simultaneously acquiring an \xray image with the marker visible in 
			both modalities. This procedure defines $t_0$ and thus 
			$^\text{C}\mat{T}_\text{W}$ in Eq.~\ref{eq:spatialRelation}. Note 
			that in the current system, a second \xray images needs to be 
			acquired for subtraction (see Sec.~\ref{subsec:marker}) but the 
			marker can now be removed from the scene.
			\item Annotate the anatomical landmarks to be targeted in the 
			fluoroscopy image as described in Sec.~\ref{subsec:hmd}.
		\end{enumerate}
		Performing the aforementioned steps yields virtual 3-D lines that may 
		provide sufficient guidance in some cases, however, the exact position 
		of the landmark on this line remains ambiguous. If the true 3-D 
		position of the landmark is needed, the above steps can be repeated for 
		another \carm pose.
		
		
		\subsection{Experiments}
		\label{subsec:experiments}
		We design experiments to separately evaluate the system's components 
		quantitatively. While the first two studies do not require user 
		interaction and objectively assess system performance, the last two 
		experiments are designed as a preliminary feasibility study that is 
		performed by two orthopaedic surgeons at the Johns Hopkins Hospital.
		
		\paragraph{Calibration} 
		In the first experiment, we seek to assess how well the \multimarker 
		enables calibration of the system. To this end, the \hmd and \carm 
		remain at fixed poses and are calibrated using the procedure described 
		in Sec.~\ref{subsec:hmd} yielding
		\begin{equation}
		^\text{C}\mat{T}_\text{HMD}(t_0) = 
		^\text{HMD}\mat{T}_\text{M}^{-1}(t_0) ~ ^\text{C}\mat{T}_\text{M}(t_0) 
		\,,
		\end{equation}
		This does not involve the world coordinate system as the pose of \hmd 
		with respect to the \carm is constant over time. Then, the \multimarker 
		is displaced multiple times and each time 
		$^\text{C}\mat{T}_\text{HMD}(t_i)$ is updated, where $i=1,\dots,6$. As 
		the spatial relation between the \carm and the \hmd remains unchanged, 
		disagreement of $^\text{C}\mat{T}_\text{HMD}(t_i)$ and 
		$^\text{C}\mat{T}_\text{HMD}(t_0)$ is related to calibration 
		performance and reproducibility. We repeat this procedure 6 times and 
		report the mean and standard deviation of positional and rotational 
		errors.
		
		\paragraph{\hmd Tracking} 
		The transformation $^\text{C}\mat{T}_\text{HMD}(t)$ is time dependent 
		as the surgeon is free to move and, thus, 
		$^\text{HMD}\mat{T}_\text{W}(t)$ changes. Accurate estimation of 
		$^\text{HMD}\mat{T}_\text{W}(t)$ is crucial to ensure that the virtual 
		objects designated for surgical guidance are displayed at the correct 
		position on the patient. To evaluate the tracking performance of the 
		\hmd, i.\,e. the \hololens, the \multimarker is fixed to the surgical 
		bed while the \hmd is mounted facing the \multimarker on a tripod to 
		avoid inaccuracies due to shaking or very fast movements. We obtain 
		reference 3-D positions of the corner points of the \multimarker in the 
		world coordinate system via:
		\begin{equation}
		^\text{M}\mat{T}_\text{W}(t_0) = ^\text{W}\mat{T}_\text{HMD}^{-1}(t_0) 
		~ ^\text{HMD}\mat{T}_\text{M}^{-1}(t_0) 	
		\end{equation}
		and then reposition the \hmd $i=1,\dots,6$ times yielding estimates of 
		$^\text{M}\mat{T}_\text{W}(t_i)$ and, thus, the 3-D corner points of 
		the \multimarker. We repeat this procedure six times and report the 
		root-mean-square error over all corner points. 
		
\begin{figure}
			\centering
			\includegraphics[width=\linewidth]{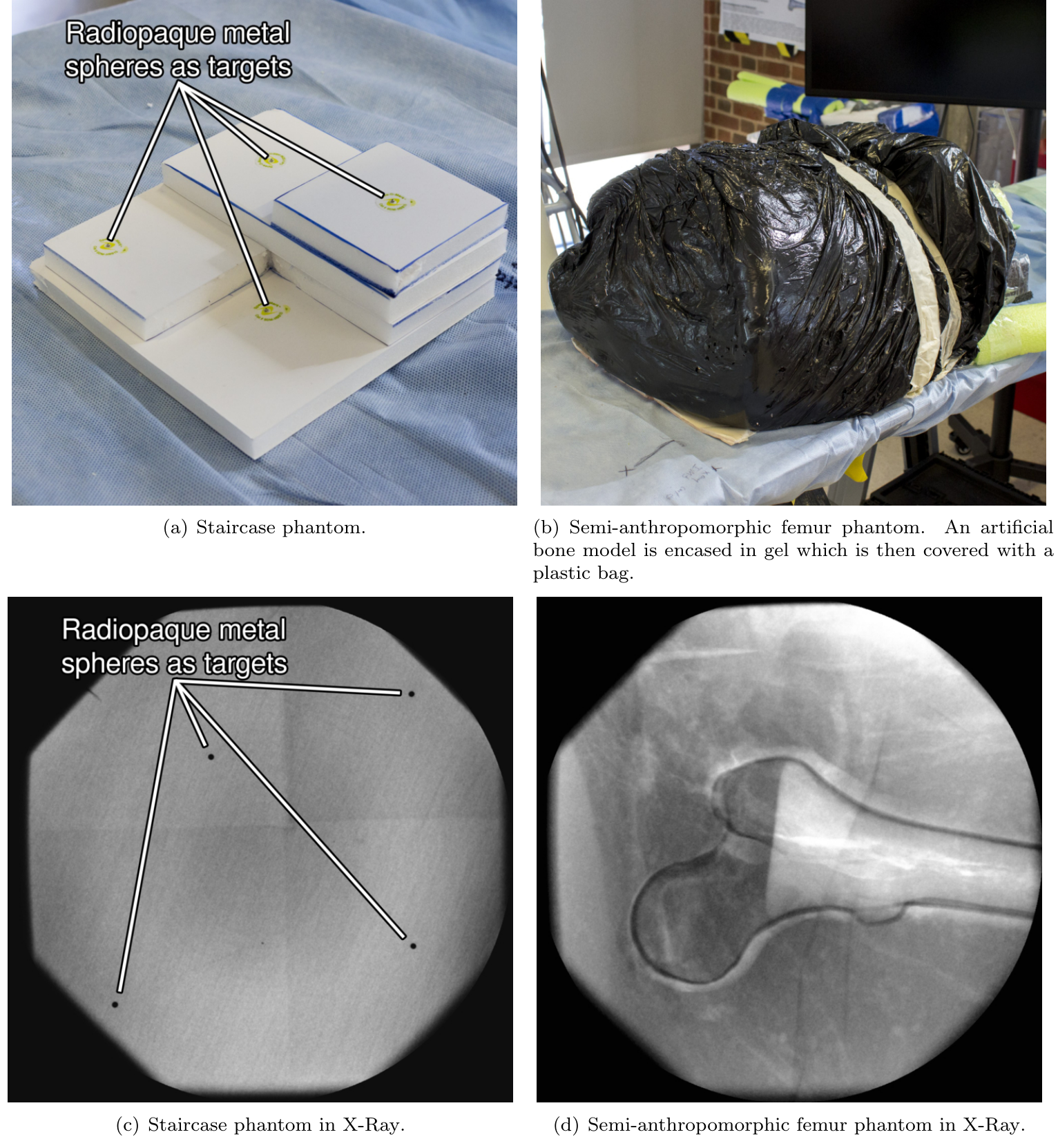}
			\caption{Phantoms used in the user studies assessing the 
			performance of the system in an isolated and a surgery-like 
			scenario in Fig.~\ref{fig:phantoms}(a) and \ref{fig:phantoms}(b), 
			and in \xray Fig.~\ref{fig:phantoms}(c) and \ref{fig:phantoms}(d) 
			respectively.}
			
			\label{fig:phantoms}
		\end{figure}

		\revision{
			\paragraph{Precision of 3-D Landmark Identification}
			We assess the precision of 3-D landmark retrieval when the system 
			is used in two-view mode. To this end, we construct a staircase 
			phantom with a metal bead attached to each plateau (see 
			Fig.~\ref{fig:phantoms}(a)). \revision{In this scenario, the metal 
			beads serve as landmarks for annotation in an otherwise featureless 
			image.} We image the phantom in two C-arm gantry positions 
			according to the workflow outlined in 
			Sec.~\ref{subsec:integration}. The first C-arm position corresponds 
			to a view where the detector is parallel to the baseplate of the 
			phantom, while the second view is rotated by approximately 
			15$^\circ$. The 3-D positions of the 4 landmark points \revis{in 
			the \carm coordinate frame} are computed via triangulation from the 
			respective corresponding annotations in the two X-ray images. The 
			experiment is repeated 5 times and the system is re-calibrated  
			using the \multimarker. To assess the precision, and thus 
			reproducibility, of the calibration, we compute the 
			\revis{centroid} and standard deviation for each of the 4 metal 
			spheres. \revis{Here, centroid refers to the mean position among 
			all corresponding landmarks.} We then state the average standard 
			deviation in millimeter as a measure for precision.
		}
		
		\paragraph{Guidance Using the AR Environment}
		The proposed system provides guidance to the surgeon by rendering a 
		virtual line that passes through the desired target that is annotated 
		in an \xray image. Assuming ideal calibration of the \carm as well as 
		perfect localization of the \hmd in the world coordinate system, it is 
		unclear how well tools, such as K-wires, can be aligned with the 
		virtual objects. This problem is emphasized as important cues such as 
		occlusion are not yet modeled for the interaction of real with virtual 
		objects.\\
		To establish an upper bound on the guidance performance, we use the 
		staircase phantom introduced above. In the user study, the phantom is 
		positioned in the center of the \carm field of view such that the base 
		is parallel to the detector plane and then covered using a layer of 
		ballistic gel. Using the \multimarker, the \carm is calibrated to the 
		\hmd and all four metal beads are annotated in an \xray image by the 
		user yielding four virtual guidance lines. The user is then asked to 
		follow each line with a K-wire penetrating the ballistic gel and 
		piercing the phantom. This targeting procedure is performed in a 
		clockwise manner 20 times (five times per marker). We then compute the 
		average distance of the punctures to the true metal bead locations.
		
		\paragraph{Semi-anthropomorphic Femur Phantom}
		\revision{Finally, we assess the complete system performance in a user 
		study on a semi-anthropomorphic femur phantom simulating entry point 
		localization for the implantation of cephalomedullary nails to treat 
		proximal femoral fractures.} The phantom is shown in 
		Fig.~\ref{fig:phantoms}(b). In order to reduce ambiguity of the desired 
		entry point, the target is defined to be the tip of the greater 
		trochanter that can be well perceived and annotated in the \xray 
		images. The phantom simulates an obese patient such that the femur is 
		encased in a thick envelope of ballistic gel. The gel casing and the 
		femur inside are positioned on the table in a manner consistent with a 
		real surgical scenario. The users are asked to navigate a K-wire onto 
		the target point first, using the conventional approach without 
		guidance and second, using the proposed system in the two view 
		scenario\revision{: As opposed to the previous experiment the 
		participants do not rely on only a single projective \xray image that 
		they can annotate, but are allowed to annotate two \xray images. 
		Annotating the same anatomical point in both images results in two 
		guidance lines intersecting at the point of interest. This point can 
		then serve as the target from an arbitrary incision point.} After the 
		K-wire is placed, its distance to the desired position is assessed in 
		3-D using cone-beam CT. Moreover, we record and report the procedure 
		time and the amount of \xray images acquired.
		
		\section{Results}
		\label{sec:results}
		
		\paragraph{Calibration}
		To measure the precision and robustness of the system calibration we 
		estimate the poses of the \multimarker from a static HMD at different 
		locations. The overall average of Euclidean distances to the 
		\revis{centroid} of measurements (positional error) between 
		$^\text{C}\mat{T}_\text{HMD}(t_i)$ and 
		$^\text{C}\mat{T}_\text{HMD}(t_0)$ is \(21.4\)\,mm with a standard 
		deviation of \(11.4\)\,mm. The overall average of angles to the average 
		orientation (rotational error) is \(0.9\)\textdegree{} with a standard 
		deviation of \(0.4\)\textdegree{}. 
		It is important to note that, as there is no ground-truth available for 
		\carm \xray poses in this experiment, we only report the consensus 
		between measurements, i.e. the precision of the calibration step.
		
		\paragraph{\hmd Tracking} 
		In the experiment evaluating the HMD tracking accuracy, we found a 
		root-mean-square error of \(16.2\)\,mm with a standard deviation of 
		\(9.5\)\,mm. As described in Section~\ref{subsec:experiments}, the 
		error is measured among all the corner points of the \multimarker.
		
		
		\revision{
			\paragraph{Precision of 3-D Landmark Identification}
			\begin{table}[h]
				\caption{Deviations of estimated 3-D landmark positions from 
				the respective \revis{centroid}. The average distance is stated 
				as a tuple of mean and standard deviation. All values are 
				stated in millimeters.}
				\centering
				\begin{tabular}{|c|c|c|c|c|c|c|}
					\hline
					& First run		& Second run		& Third run		& 
					Fourth run	& Fifth run	& Average \\
					\hline
					\hline
					Target P1		& 19.3	& 3.54	& 4.78	& 8.31	& 11.6 	& 
					\((9.49, 	6.31)\) \\
					Target P2		& 12.9	& 4.22	& 7.67	& 11.5	& 7.56	& 
					\((8.76, 	3.44)\) \\
					Target P3		& 18.2 	& 4.19	& 6.76	& 10.4	& 6.29 	& 
					\((9.18, 	5.53)\) \\
					Target P4		& 21.6	& 6.62	& 3.51	& 18.8	& 8.23 	& 
					\((11.7, 	7.93)\) \\
					\hline
				\end{tabular}
				\label{tab:errorxyzStepPhantom}
				
			\end{table}
			
			As stated in Table~\ref{tab:errorxyzStepPhantom}, the average 
			distance to the \revis{centroids} ranged from \(8.76\)\,mm to 
			\(11.7\)\,mm. The in-plane error, evaluated by projecting the 3-D 
			deviations onto the detector plane of the first \xray orientation 
			is substantially lower and ranges from \(3.21\)\,mm to 
			\(4.03\)\,mm, as summarized in Table~\ref{tab:errorxzStepPhantom}. 
			
			\begin{table}[h]
				\caption{Deviations of estimated 3-D landmark positions from 
				the respective \revis{centroid} projected onto the \xray plane 
				of the first view. Again, the average distance is stated as a 
				tuple of mean and standard deviation and all values are given 
				in millimeters.}
				\centering
				\begin{tabular}{|c|c|c|c|c|c|c|}
					\hline
					Distance to  & First run		& Second run		& Third 
					run		& Fourth run	& Fifth run	& Average \\
					\hline
					\hline
					\revis{Centroid} P1		& 4.59	& 3.42	& 4.78	& 0.32	& 
					2.96	& \((3.21, 	1.79)\) \\
					\revis{Centroid} P2		& 5.05	& 3.71	& 6.06	& 3.17	& 
					0.36	& \((3.67, 	2.17)\) \\
					\revis{Centroid} P3		& 3.93 	& 3.82	& 5.11	& 3.32	& 
					3.57 	& \((3.96, 	0.70)\) \\
					\revis{Centroid} P4		& 6.48	& 4.24	& 3.45	& 4.69	& 
					1.31 	& \((4.03, 	1.89)\) \\
					\hline
				\end{tabular}
				\label{tab:errorxzStepPhantom}
			\end{table}
		}
		
\begin{figure}
			\centering
			\includegraphics[width=\linewidth]{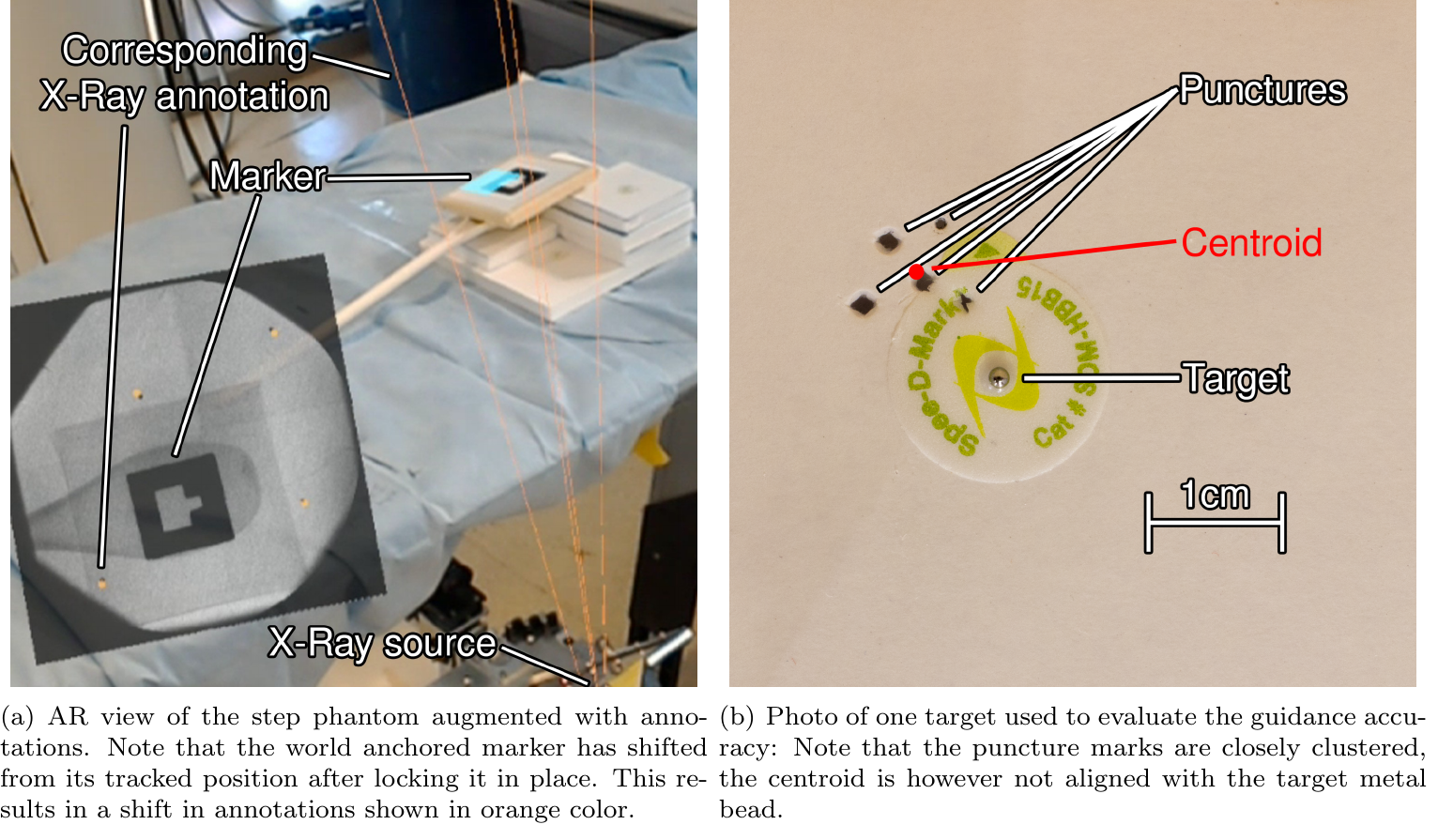}
			\caption{Setup and the AR view for the "Guidance Experiment Using 
			the AR Environment".}
			\label{fig:arSetup}
		\end{figure}

		\paragraph{Guidance Using the AR Environment}
		\revision{The experiment performed by two expert users on step phantoms 
		as shown in Fig.~\ref{fig:phantoms}(a) resulted in an average precision 
		error of \(4.47\)\,mm with a standard deviation of \(2.91\)\,mm 
		measured as the average Euclidean distance to the \revis{centroid} of 
		the puncture marks.} The accuracy of this system is then measured as 
		the average distance to metal beads i.e. ground-truth, which yielded 
		\(9.84\)\,mm error with a standard deviation of \(3.97\)\,mm. 
		\revision{The mean errors and standard deviations of both participants 
		are summarized in Table~\ref{tab:meanstdStepPhantom}, showing the 
		average distance of each attempt to the \revis{centroid} of all 
		attempts and the average distance of this \revis{centroid} to the true 
		metal bead target.} Fig.~\ref{fig:arSetup}(b) shows the target and the 
		\revis{centroid} on the step phantom. 
		
		\begin{table}
			\caption{Errors measured as distances to the target and the 
			\revis{centroid}. All values are stated in millimeters in tuples of 
			mean and standard deviation. Results are shown for both 
			participants (P1 and P2).}
			\centering
			\begin{tabular}{|r|c|c|c|c|c|}
				\hline
				~ Distance to	& First target		& Second target		& Third 
				target		& Fourth target		& Overall \\
				\hline
				Target P1		& \((8.17, 3.85)\)	& \((8.08, 1.63)\)	& 
				\((7.97, 4.08)\)	& \((5.65, 1.56)\)	& \((7.47, 3.20)\) \\
				\revis{Centroid} P1		& \((5.93, 4.02)\)	& \((2.49, 1.29)\)	
				& \((5.84, 2.40)\)	& \((2.49, 1.22)\)	& \((4.19, 3.03)\) \\
				Target P2		& \((12.3, 2.1)\)	& \((11.3, 0.8)\)	& 
				\((15.2, 3.9)\)	& \((10.1, 2.4)\)	& \((12.2, 3.2)\) \\
				\revis{Centroid} P2		& \((4.92, 2.96)\)	& \((4.54, 2.80)\)	
				& \((6.30, 2.80)\)	& \((3.25, 1.27)\)	& \((4.75, 2.77)\) \\
				\hline
			\end{tabular}
			\label{tab:meanstdStepPhantom}
			
		\end{table}
		
\begin{figure}
			\centering
			\includegraphics[width=0.5\linewidth]{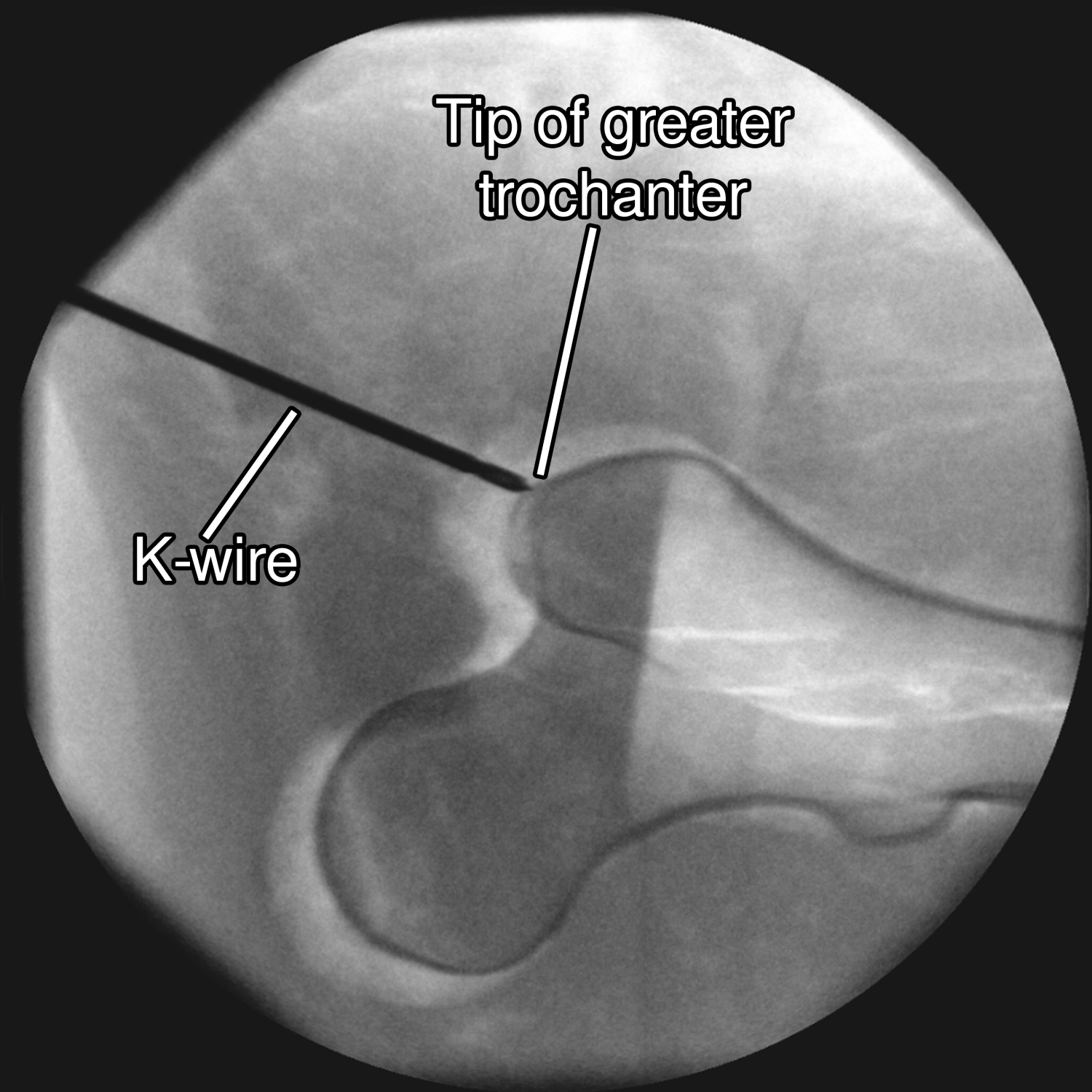}
			\caption{\xray of the \revision{second experiment on a} 
			semi-anthropomorphic Femur Phantom with the K-wire in the final 
			position for one of the participants.}
			\label{fig:berthapoked}
		\end{figure}

		\paragraph{Semi-anthropomorphic Femur Phantom} For this experiment, 
		surgeons were asked to perform a simulated K-wire placement on the 
		semi-anthropomorphic femur phantom with the on-the-fly AR system, as 
		well as classic fluoro-guided approach. The average distance of the tip 
		of the K-wire to the tip of the greater trochanter is \(5.20\)\,mm with 
		the proposed AR solution and \(4.60\)\,mm when only fluoroscopic images 
		were used. However, when the proposed solution was used, the average 
		number of \xray images substantially decreased. The participants needed 
		5 \xray acquisitions from 2 orientations and on average 16 \xray images 
		from 6 orientations when the proposed and traditional solution were 
		used, respectively. In fact, the number of images for our solution can 
		further be decreased by two as we include the images required for 
		background subtraction that may become obsolete with a different marker 
		design. 
		Finally, the procedure time reduced from $186\,$s (standard deviation 
		of $5$\,s) in the classic approach to $168\,$s (standard deviation of 
		$18$\,s). 
		An \xray of the K-wire at the final position of one of the participants 
		can be seen in \ref{fig:berthapoked}.

		\section{Discussion}
		\label{sec:discussion}
		
		\subsection{Outcome of the Preliminary Feasibility Studies}
		\label{subsec:discPrelim}
		The experiments conducted in this paper are designed to distinguish 
		between accuracy and precision errors of three different parts of the 
		proposed AR support system: the calibration between the RGB camera of 
		the \hmd and the \carm, the world tracking of the \hmd, and the 
		visualization of the guidance.\\
		The results of the calibration experiment, where the C-arm source is 
		tracked with respect to the \hmd, indicate large positional error and 
		low orientational error. Two main sources for this error are 
		\textit{i)} error propagation due to large distances between the HMD, 
		marker, and the X-ray source where small errors in marker tracking 
		translate to large displacements in the estimation of the pose of the 
		X-ray source, and \textit{ii)} errors in marker tracking that increase 
		when the \multimarker is not facing parallel to the RGB camera on the 
		\hmd. The \textit{HMD Tracking} experiment indicates a drift in 
		tracking of the \multimarker with respect to the world anchor as the 
		user observes the marker from different locations during the 
		intervention. This error decreases in a static environment where the 
		spatial map of the \hmd works more reliably.
		\revision{All aforementioned sources of error affect the 
		reproducibility reported in the \textit{3-D landmark identification} 
		experiment. The distance from the \revis{centroid} was substantially 
		reduced when only the in-plane error was considered, an observation 
		that is well explained by the narrow baseline between the two \xray 
		poses. Yet, the in-plane distance found in this experiment is in good 
		agreement with the precision reported for the user study on a similar 
		step-phantom.}
		\\
		\revision{The quantitative error measures reported in 
		Section~\ref{sec:results} suggest lackluster performance of some of the 
		subsystem components that would inhibit clinical deployment for 
		procedures where very high accuracy is paramount. However, in scenarios 
		where rough guidance is acceptable, the overall system performance 
		evaluated on the semi-anthropomorphic femur phantom is promising. The 
		distance of the K-wire from the anatomical landmark is comparable, yet, 
		the proposed system requires the acquisition of fewer \xray images. The 
		results suggest that the proposed on-the-fly AR solution may already be 
		adequate to support surgeons in bringing surgical instruments close to 
		the desired anatomical landmarks. It is worth mentioning that many of 
		the limitations discussed here are imposed on our prototype solution as 
		it relies on currently available hardware or software. Consequently, 
		improvements in these devices will directly benefit the proposed 
		workflow for on-the-fly AR in surgery.
		}
		
		\subsection{Challenges}
		
		Our method combines two approaches to create an easy guidance system.
		First, it utilizes the accessible tracking capability of the \hmd, to 
		use the spatial map and its world anchor as the fundamental coordinate 
		system of the tracking.
		Likewise the AR feature is used as a straightforward visualization 
		technique, guiding the surgeons without any external tool tracking but 
		allowing them to do the final registration step between surgical tool 
		and guidance system intuitively by themselves.\\
		Calibration using the proposed \multimarker is straightforward and 
		proved to be reasonably accurate considering the current design. Use of 
		the marker for calibration of the \carm to the \hmd is a convenient 
		solution due to its flexibility that is slightly impeded by the need 
		for prior offline calibration of the intrinsic parameters. 
		\revision{Consequently, it would be beneficial to investigate other 
		marker designs that would enable simultaneous calibration of intrinsic 
		and extrinsic parameters of all cameras and thus promote the 
		ease-of-use even further. In the same line of reasoning, although 
		\artoolkit is a well known tool for camera calibration via marker 
		tracking, it might not be the best solution here as it is not designed 
		for transmission imaging which partly explains the low accuracy 
		reported in the calibration experiment. 
			The marker design itself, i.\,e. the sheet of lead, imposes the 
			need for digital subtraction, which increases the required number 
			of \xray images and, thus, the dose by a factor of two, which may 
			not be favorable in the clinical scenario. However, the need for 
			subtraction is conditional on the design of the marker. Within this 
			feasibility study we valued convenient processing using \artoolkit 
			over dose reduction and thus required subtraction imaging. This 
			requirement, however may become obsolete when transitioning to more 
			advanced marker designs, e.\,g. by combining \artoolkit markers 
			with small metal spheres for RGB and \xray calibration, 
			respectively. Such advanced approaches would further allow 
			simultaneous calibration of both extrinsic and intrinsic 
			parameters.\\}
		While the localization of the \hmd with respect to the world coordinate 
		system works well in most cases, it proved unreliable in scenarios 
		where the surroundings are unknown, i.\,e. at the beginning of the 
		procedure, or in presence of large changes in the environment, such as 
		moving persons. While this shortcoming does affect the quantitative 
		results reported here, it does not impair the relevance of the proposed 
		guidance solution as novel, more powerful devices and algorithms for 
		SLAM will become available in the future.\\
		The \hmd used here adjusts the rendering of the virtual objects based 
		on the interpupillary distance~\cite{Hollister2017}. However, despite 
		its name and advertisement, the \hololens is not a holographic display 
		as all virtual objects are rendered at a focal distance of 
		approximately 2\,m. The depth cue of accommodation is thus not 
		available. This is particularly problematic for unexperienced users, as 
		the K-wire on the patient and the virtual objects designated for 
		guidance cannot be perceived in focus at the same time. Moreover, there 
		is currently no mechanism available that allows for a natural 
		interaction between real and virtual objects. Consequently, cues that 
		enable correct alignment of the tools with the guidance line, such as 
		accurate occlusions or shading, cannot be provided. The difficulties in 
		alignment are reflected in the staircase phantom of the feasibility 
		studies. While these shortcoming affect the current prototype, we 
		expect these challenges to be mitigated in the future when more 
		sophisticated AR hardware becomes available.

		\subsection{Limitations}
		The proposed system required a wireless data sharing network to stream 
		intra-operative images to the \hmd. While this requirement may be 
		considered a drawback at this very moment, it may be seen as less 
		unfavorable when intra-operative inspection of medical images 
		transitions from traditional to virtual 
		monitors~\cite{chimenti2015google,yoon2016technical,qian2017technical}. 
		Conceptually it may even be possible to perform this "on-the-fly" 
		guidance without any connection to the \carm or other additional 
		surgical hardware by using only the \hmd and the \multimarker. In this 
		version, the radiographic projection of the fiducial is directly 
		observed on the physical monitor using the RGB camera of the \hmd. 
		Annotations can then be made directly on an augmented reality plane 
		that overlies the radiographic image at the position of the physical 
		monitor, potentially allowing for AR guidance in completely unprepared 
		environments. However, use of the radiology monitor rather than the raw 
		\xray image introduces substantial additional sources of error into the 
		system, as the pose of the \hmd with respect to the monitor plane must 
		be very accurately known. Furthermore, the ability to annotate the 
		image with this method could be difficult depending on the position of 
		the monitor in the room and the distance between the surgeon and the 
		monitor.\\
		Wearing an \hmd may feel uncomfortable to some surgeons. We hypothesize 
		that the need to wear an \hmd during surgery is not a major impediment 
		for orthopaedic surgery, where head based tools, such as magnification 
		loupes, sterile surgical helmets, and headlamps, which are heavier than 
		an \hmd and many times require tethering, are already part of clinical 
		routine.

		\section{Conclusion}
		\label{sec:conclusion}
		
		We proposed an easy-to-use guidance system for orthopaedic surgery that 
		co-calibrates a \carm system to an optical see-through \hmd to enable 
		on-the-fly AR in minimally prepared environments. Co-calibration of the 
		devices is achieved using a \multimarker that are then registered to 
		the world coordinate frame using the SLAM tracking of the \hmd. After 
		calibration, point and line annotations in the 2D \xray images are 
		rendered using the \hmd as the corresponding virtual lines and planes 
		in 3D space, respectively, that serve as guidance to the surgeon.\\
		The performance of the proposed system is promising for starting point 
		localization in percutaneous procedures, and could benefit from future 
		advances of AR. Particularly, future work should consider possibilities 
		to improve on the interaction of real and virtual objects, as the 
		current lack of depth cues impedes superior performance. 

		\subsection*{Disclosures}
		The authors declare that they have no conflict of interest.\\
		\noindent Research reported in this publication was partially supported 
		by Johns Hopkins University internal funding sources.
		
		\acknowledgments 
		The authors want to thank Gerhard Kleinzig and Sebastian Vogt from 
		Siemens Healthineers for their support and making a Siemens ARCADIS 
		Orbic 3D available.


	\end{spacing}
\end{document}